\documentclass[journal]{IEEEtran}

\usepackage{cite}
\usepackage{amsmath}
\usepackage{amsthm}
\usepackage{amsfonts}
\usepackage{verbatim}
\usepackage{color}
\usepackage{graphicx}
\usepackage{subfigure}
\usepackage{algorithmic,algorithm}
\usepackage{enumerate}
\usepackage{epstopdf}
\usepackage{multirow}
\usepackage{color}

\newcommand{\figdir}{figures}
\def\swidthone{0.75\linewidth}
\def\swidthtwo{0.45\linewidth}

\newcommand{\argmax}{\operatornamewithlimits{argmax}}
\newcommand{\argmin}{\operatornamewithlimits{argmin}}
\begin{document}

\title{Microscopic Muscle Image Enhancement}

\author{Xiangfei~Kong,~\IEEEmembership{Member,~IEEE,}
        and~Lin~Yang,~\IEEEmembership{Member,~IEEE}
\thanks{X. Kong and L. Yang are with the J. Crayton Pruitt Family department
of Biomedical engineering, University of Florida, Gainesville, FL, 32611 USA.
E-mail:Xiangfei.Kong@bme.ufl.edu}
}

\markboth{Journal of XXXX,~Vol.~XX, No.~X, Jue~2015}%
{Kong \MakeLowercase{\textit{\emph{et al.}}}: Microscopic Muscle Image Enhancement}

\maketitle

\begin{abstract}
We propose a robust image enhancement algorithm dedicated for muscle fiber specimen images captured by optical microscopes. Blur or out of focus problems are prevalent in muscle images during the image acquisition stage. Traditional image deconvolution methods do not work since they assume the blur kernels are known and also produce ring artifacts.  We provide a compact framework which involves a novel spatially non-uniform blind deblurring approach specialized to muscle images which automatically detects and alleviates degraded regions. Ring artifacts problems are addressed and a kernel propagation strategy is proposed to speedup the algorithm and deals with the high non-uniformity of the blur kernels on muscle images. Experiments show that the proposed framework performs well on muscle images taken with modern advanced optical microscopes. Our framework is free of laborious parameter settings and is computationally efficient.
\end{abstract}


%
\IEEEpeerreviewmaketitle

\section{Introduction}
\IEEEPARstart{S}{keletal} muscle is an extremely adaptive tissue that is able to change size depending upon external stimuli. Increases in muscle mass in response to resistance training and losses in mass caused by disuse or associated with chronic diseases such as cancer and HIV are primarily due to hypertrophy or atrophy of individual muscle fibers, respectively, rather than addition or loss of fibers \cite{Muscle_fiber}. Thus, The ability to accurately and efficiently quantify the morphological characteristics of muscle cells, such as cross-sectional areas (\textbf{CSA}s), is vital for assessing muscle function, since muscle mass is the primary determinant of muscle strength \cite{Muscle_strength1,Muscle_strength2}. One example is the Idiopathic Inflammatory Myophathy (IIM) detection, which is characterized by weakness and inflammation of skeletal muscles \cite{Muscle_IIM}. A very important prerequisite for accurate CSA quantification is the high-quality microscopic image acquisition, after the muscle specimen is stained with H\&E or fluorescence.

Unfortunately, the non-flat surface of muscle specimens, which involves massive tissue protuberances and concavities, impedes the users holding the desired observatory regions to uniformly fall into the focus plane of the microscope. Nevertheless, modern advanced optical microscopes for fluorescence or H\&E specimens often involve a lot of laborious manual configurations of confusing parameters. It is usually a prerequisite for the users to be well trained and extensively practiced in order to properly adjust those settings. Thus, improper or even erroneous settings of parameters are inevitable. Blur/defocus is consequently produced occasionally at some regions, or the entire scope of the captured muscle microscopic images. Vagueness and ambiguity are observed on these regions not only degrading the visual quality but also hindering the success of the subsequent computer aided diagnosis operations potentially included in an automatic diagnosis system such as segmentation of cells and the detection of a certain disease. 

In this work, we propose a compact framework to enhance the visual quality of the poorly captured muscle images. The proposed framework involves the following steps: 1) Local blur/defocus kernel estimation. A group of spatially variant blur kernels for part of the muscle image are estimated based on the gradient information under carefully chosen constraints and priors. 2) Kernel propagation. The estimated kernels are propagated to their neighborhood and speedup their kernel estimation. 3) Non-blind image enhancement. An efficient image deconvolution algorithm involving the most recent findings in image statistics is performed locally based on the estimated kernels. Fast convergence speed is achieved. The proposed method has an automatic design that reduces blur and out of focus in a one-click style. The users are thus liberated from laborious parameter tweaks.

\begin{figure}[t]
	\begin{center}
		\begin{tabular}{cc}
			\includegraphics[width=\swidthtwo]{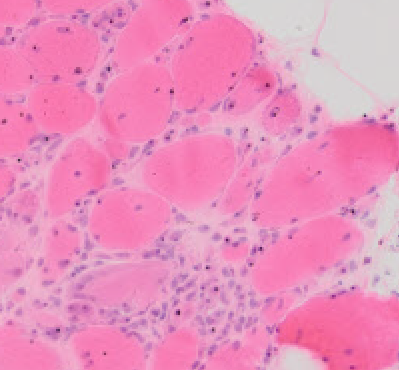}&
			\includegraphics[width=\swidthtwo]{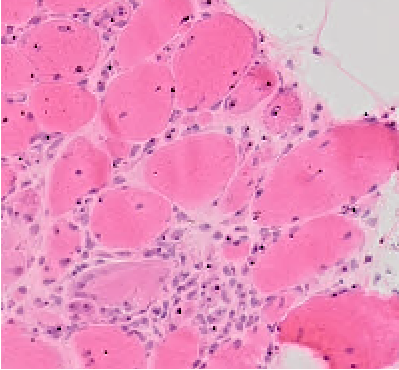}\\
			(a) Blurry & (b) Enhanced\\
			\includegraphics[width=\swidthtwo]{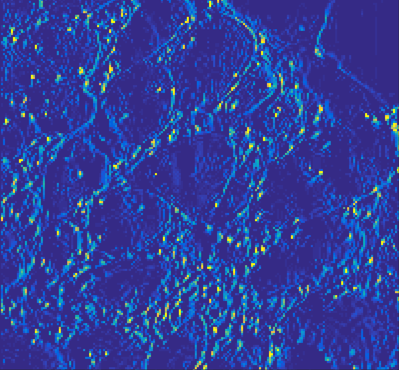}&
			\includegraphics[width=\swidthtwo]{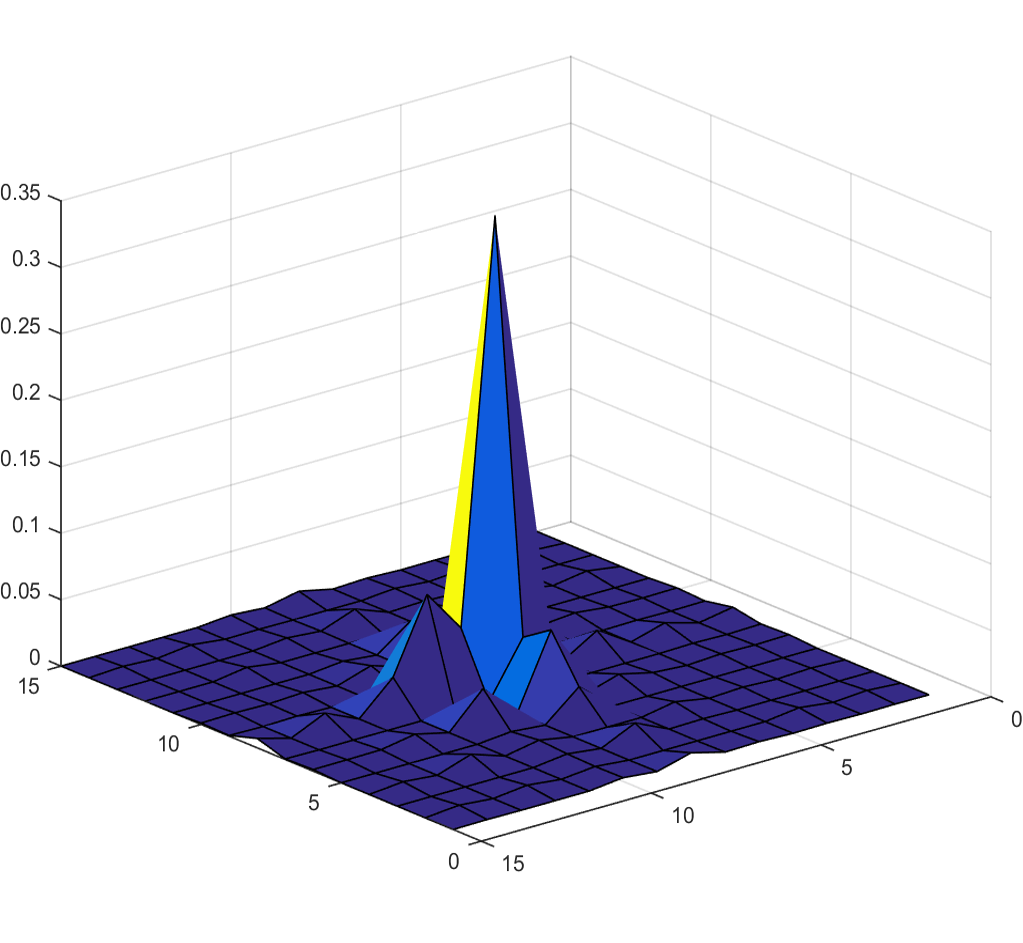}\\
			(c) Gradient & (d) Kernel
		\end{tabular}
	\end{center}
	\caption{An example of (a) a blur/defocus region observed on a microscopic muscle images and (b) enhanced by the proposed framework. (c) The gradient map of (a) used in the proposed method. Darker color corresponds to lower values. (d) The estimated blur kernel. 
	}
	\label{Fig:example}
\end{figure}

We do not confine the type of the degradation as specific defocus caused by the limited or misplaced depth of field (DoF) produced by the optical system of the microscope. Rather, we put it into a more general blur model that captures the local linear transforms represented as a (or a group of) freely shaped kernel to cope with the degradation caused by sophisticated blur such as lens aberration \cite{Aberration_nature}.
The proposed muscle image enhancement is characterized as blind, non-uniform blur kernel estimation and deconvolution with the purposes of both computational efficiency and automation. Conventional methods, such as as Weiner filters \cite{WeinerFilter} and Richard-Lucy deconvolution \cite{Deblur_Lucy}, are parametric since they assume the precise blur kernels are known. In addition, the enhanced images provided by them are often less constrained and thus have strong artifacts. The natural image defocus removal methods such as \cite{Defocus} usually rely on a robust map estimation of Gaussian parameter and the existence of similar patch pairs between the blurry and the good regions. In contrast, the proposed method utilizes the sparsity of the gradient as clue to guide the algorithm automatically produce good quality results. No heuristic computations are involved and the noise is also addressed with proper models. The enhancement framework is thus both robust and automatic.

The blur kernel estimation step involved in the proposed framework differs from the existing non-uniform methods in literature significantly. Those works are dedicated to model the spatial motion of a hand-held camera. 3D translations, in-plane and/or out-of-plane rotations are involved and are estimated during the energy optimization stage. However, rotations do not happen on a microscopic muscle image. What is more important is that the blur happened in microscopic muscle images is much different from typical natural image blur, exhibiting much more severe spatial variances but less (still exists though) directional blur. Although part of the blur can be understood as out of focus and is proper to be modeled using an exponential family function (such as Gaussian or Poisson), putting such rigid constraints does not perform well for lack of flexibility, especially on muscle images with strong spatial variance. We address this problem by proposing a new spatially variant kernel estimation algorithm. Specifically, a group of spatially invariant kernels are estimated for some randomly chosen local image regions. These estimations are propagated to their neighboring regions serving as initiations of their optimization arguments. This step not only preserves the coherence of the inter-region image structures but also accelerates the computational efficiency drastically. Although we also encourage the estimated kernels to be Gaussian-shaped, the weight assigned to this prior is gradually deducted during the iteration of the algorithm. This brings the benefit that the blur is treated as defocus problem at the first few iterations to produce a reasonable intermediate result. The most recent findings in image statistics suggesting the sparsity constraint on gradient of the image are utilized during the kernel estimation. An efficient deconvolution method is used to recover the blurred muscle images after all kernels are estimated.

Although the proposed method can effectively remove the blur and defocus, the recovery of some regions that are severely blurred or defocused cannot rely on only local neighbored pixel information that typical deblur methods utilize since the original image information is mostly lost. We propose to use a patch synthesis algorithm to address this problem. These severely blurred regions are detected via a simple gradient computation after the muscle image is processed via the proposed spatially non-uniform deblur algorithm. Then, a fast non-local similar patch search is used to find out candidates of synthesis for the detected regions. Candidates with both high similarity and good patch-wise coherence preservation are selected and replace the highly degraded patches. Finally, a smoothing algorithm is performed to remove the artifacts produced by during the synthesis step.

Extensive experiments are conducted on muscle images captured with advanced optical microscopes where both mild and severe blur happen. We use an Olympus VS 120 advanced optical microscope dedicated to medical specimen observation and capture 44 muscle images. Most images more or less contain some regions showing blur or defocus problems. One example of the enhancement provided by our framework is shown in Figure \ref{Fig:example}. The users are free of empirically choosing parameters for our algorithm since the framework is completely automatic. The processing time for each three color channel muscle image ranges from two minutes to eleven minutes, depending on the resolution which is normally more than 10m pixels. Results show that the proposed framework is both effective and computationally efficient.

The remaining of this paper is organized as follows: We briefly survey relevant works in the current literature for the proposed framework in section \ref{sec:relatedWork}. The proposed muscle image enhancement framework is introduced in details in section \ref{sec:deblur}. We present the experimental results in section \ref{sec:experiment}. We conclude our paper and give a discussion on the future work in section \ref{sec:conclusion}.

\section{Related Work}
\label{sec:relatedWork}
The most relevant works to the proposed muscle image enhancement framework can be categorized as the single image blind deconvolution. The problem is typically modeled as a blur kernel (or point spread function (\textbf{PSF})) convolves with the latent, distortion free image. The blur kernel can be a Gaussian-shaped distribution to model out of focuse (defocus), or a motion trajectory to model the blur caused by spatial shifts of a camera during a long time exposure. Blind image deconvolution is considered as extremely difficult due to its severe ill-posedness since both the latent image and the blur kernel are assumed unknown. Traditional methods such as Weiner filters \cite{WeinerFilter} and Richard-Lucy deconvolution \cite{Deblur_Lucy} are popular due to their simplicity and efficiency. However, these methods cannot work on blind image deconvolution since they assume the blur kernel to the latent image is known. In addition, these works produce large amount of undesired ring-like artifacts during the latent image recovery, even if we assume the kernel is available (hand-designed model with super-tweaked parameters). In the past decade, there has been a significant progress on the blind image deconvolution involving both the phases of blur kernel estimation and latent image recovery. Various constraints are designed on both the latent image and the blur kernel with much more improved results being obtained. The reason of their success is twofold: 1) The constraints made on the kernel and the latent image alleviate the ill-posedness of the blind image deconvolution problem. 2) Well designed constraints can effectively suppress the artifacts on the latent image and exclude those unnatural latent images from the solution space. Noise in the estimated blur kernel can be also reduced with a proper constraint on it. The former produces more visually pleasant result while the latter usually leads to more robust kernel estimations. 

Shan \emph{et al.} apply the recent findings derived from image statistics that in spatial domain, the natural image gradient should follow a heavy-tailed distribution \cite{imageStatistics} and make such constraint on their latent image in gradient domain \cite{Deblur_ShanQi}. Goldstein \emph{et al.} use the power law of natural image in frequency domain
\begin{equation*}
|\hat{I}(\omega)^2|\propto ||\omega||^{-\beta},
\end{equation*}
where $\hat{I}$ is the Fourier transform of a natural image, $\omega$ is the frequency variable and $\beta$ is assumed to be roughly two \cite{Deblur_PowerSpectrum1}. Yue \emph{et al.} further exploit the power spectrum of the natural images and propose a new model for the blur kernel in frequency domain \cite{Deblur_PowerSpectrum2}.
However, the image statistics of both spatial and spectrum domain may not hold for muscle images since they are significantly different from the natural images and often contain more special shape and texture pattern. Weak edges, subtle curves or corners included in muscle images usually convey important pathological meanings and diagnostic clues. The sparsity on the gradient map for the latent image is a more general and robust regularization promising for muscle images. In this context, Whyte \emph{et al.} adopt $l_1$ sparsity to constrain the gradient \cite{Deblur_Whyte} of the latent image. Krishnan \emph{et al.} further improve this term by normalizing the $l_1$ norm with the $l_2$ norm to better express the nature of the blur \cite{Deblur_Krishnan}. Xu et. al adopt non-linear penalties to $l_1$ norm \cite{Deblur_XuLi2} and sophisticated $l_0$ norm \cite{Deblur_XuLi_l0} \cite{Deblur_XuLi1} of the gradient of the latent image. The $l_1$ norm \cite{Deblur_Krishnan} \cite{Deblur_HuZhe} or the $l_2$ norm \cite{Deblur_Cho} are typically used to regularize the estimated kernel.

Without proper constraints, ring artifacts are usually produced due to outliers/noise of the data or the saturated pixels. Shan et. al propose a heuristic method for ring effect reduction by localizing the smooth regions of the latent image and suppress their gradients \cite{Deblur_ShanQi}. Whyte \emph{et al.} examine the reasons why ring artifacts emerge and propose that it is because of the nonlinear mapping caused by saturated pixels and the lack of constrains when solving on the near-zero domain of the transform made by the kernel \cite{Deblur_Whyte}. They propose to use a mask to filter out those saturated pixels to avoid wrong estimations and also use some regularizations to further forbid those errors from propagating to good estimations. Cho \emph{et al.} propose a method in a non-blind deconvolution with similar rationale but the mask and the outliers are explicitly modeled and optimized, leading to a better result \cite{Deconvolution_Cho}.

The context of aforementioned deblur algorithms is spatially uniform which assumes a constant kernel across the entire image. Despite their amazingly good performance \cite{Deblur_benckmark}, this is not true in either natural images taken with a camera or the microscopic muscle images in this paper. Several works \cite{Deblur_nonUniform_Gupa} \cite{Deconvolution_l1RL_MBrown} \cite{Deblur_XuLi1} extend their kernel models to 3D or 6D degree of freedom (\textit{DoM})%
\footnote{Suppose the 2D image locates on the plane determined by the $x$ and $z$ axises, the 3D modeling is either rotations along the three axises $\{\theta_x,\theta_y,\theta_z\}$, or the translations on $x$ and $z$ axises and the rotation on visual axis $z$ $\{\tau_x,\tau_y,\theta_y\}$. The 6D modeling is to use both. See \cite{Deblur_benckmark} for more details.}, 
describing a complete motion domain caused by in-plane translations, rotations, and out-plane rotations.  However, the performance of the deblur does improve significantly \cite{Deblur_benckmark}. We show in this paper that uniform assumption is good for local regions of the muscle image while non-uniformity must be considered in the global domain. 

\section{Muscle Image Enhancement}
\label{sec:deblur}
\begin{figure*}
\begin{centering}
	\includegraphics[width=\swidthone]{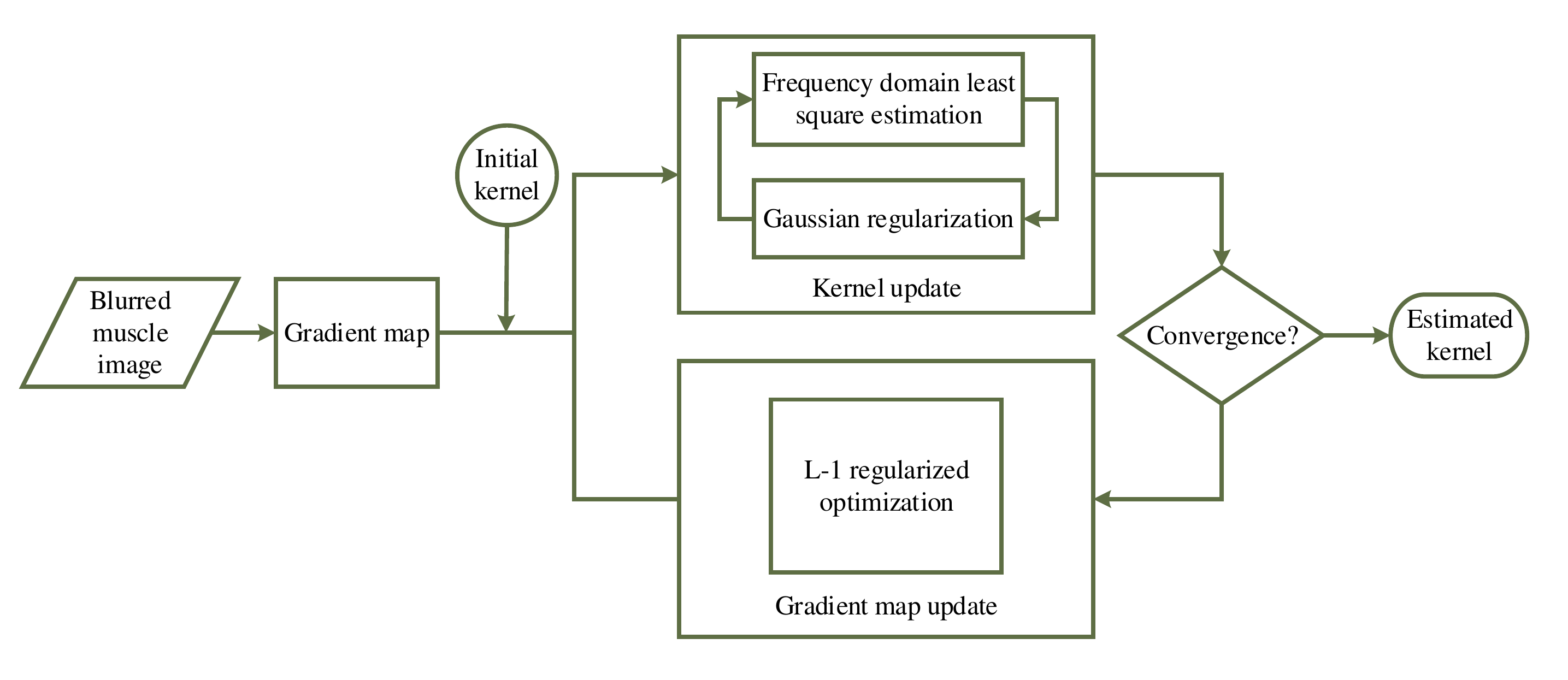}
	\caption{The flowchart of the proposed method.}
	\label{Fig:flowchart}
\end{centering}
\end{figure*}

In this section, we introduce the proposed spatially non-uniform deblur framework involving blur kernel estimation and muscle image deconvolution. We do not assume the blur kernel for muscle images is known and explicitly estimate it. The proposed kernel estimation algorithm is overall spatially variant but can be separated into a group of spatially invariant estimations. These estimations are made on a few random subregions of the muscle image. Estimated kernels are treated as seeds and are propagated to their neighboring regions as the initializations for kernel estimation. We show in the experimental results that the proposed scheme has fast convergence maintains good coherence of the image structure. An overall flowchart of the proposed method is shown in Fig. \ref{Fig:flowchart}.
\subsection{Kernel Estimation}
In the spatially invariant blur kernel estimation context, one effective Bayesian inference approach, namely, the MAP model, is frequently used. Suppose the latent image (or a subregion) $X$ is blurred with a kernel $k$ with some additive noise $\epsilon$ that follows a specific statistical distribution. The degradation of blur and defocus on a latent image can be both modeled as convolution computation written as:

\begin{equation}\label{Eq:deblur_overall}
I=X\otimes k+\epsilon,
\end{equation}
where $I$ denotes the degraded image, $X$ is the latent distortion-free image, and $\otimes$ denotes convolution operation. Note that there is an option to reorganize $k$ or $X$ to write the convolution as one matrix multiplies the other as in Eq. \ref{Eq:deblur_overall_v2}:

\begin{equation}\label{Eq:deblur_overall_v2}
I = AX + \epsilon,
\end{equation}
where $A$ is the Toeplitz form of the kernel $k$\footnote{Each row of $A$ corresponds to a vectorized version of $k$ with a phase shifted by one element (and reversed for convolution).}. However, this is not necessary for our algorithm and all terms are written as their original version. The blur kernel $k$, or PSF, can be either a motion trajectory to model the physical motion blur, or a Gaussian distribution to model defocus, or both (in which the kernel usually shows a blurred version of a motion trajectory). In MAP-estimation framework, deblur algorithms typically assume the pixels of the image are independent and attempt to maximize the posteriori $p(X,k|I)$:

\begin{equation}
\max_{X,k} p(X,k|I)\propto \prod^{N}_{i=1} p(I|x_i,k)P(k)p(X),
\end{equation}
where $p(I|x_i,k)$ is the likelihood of the degraded image and $X=\{x_i|i=1...N\}$ where $N$ denotes the number of pixels of the latent image. $p(k)$, $p(x)$ are the prior probability of the kernel and the latent image, respectively. This form can be interpreted as an energy minimization problem by defining $E(X,k)=-log*(p(X,k|k))$ and write Eq. \ref{Eq:deblur_overall} as:

\begin{equation}
\min_{X,k} E(X,k) \propto \sum^N_{i=1} E(I|x_i,k)+E(k)+E(X),
\end{equation}
where $\sum^N_{i=1}E(I|x_i,k) $ is referred to as the data fidelity term. $E(k)$ and $E(X)$ are the priors of the blur kernel and the latent image, respectively. In the following sections, we introduce the choices of the MAP terms in the proposed muscle image enhancement framework and provide the optimization solution.

\subsubsection{Data Fidelity and Noise Model}
The purpose of the data fidelity term is to constraint the estimated latent image not to deviate too much from its blurred version. Recent findings show that gradient plays an essential role during the kernel estimation \cite{Deblur_ShanQi}. In fact, $X$ is not necessarily a ``real" image but can be either gradient map $(\partial_xx_i,\partial_yx_i)$ \cite{Deblur_ShanQi} \cite{Deblur_Krishnan} or some ``unnatural" representations of the image \cite{Deblur_Cho} \cite{Deblur_XuLi1}. Due to the severe ill-posed nature of this problem, these options usually provide more helpful information of the blur kernel than the natural image estimated during the optimization or the initial blurred image. After the MAP optimization reaches the convergence, only the estimated kernel is used and the final output image is computed using some non-blind deconvolution approaches with  image priors to suppress artifacts. In this paper, we propose to use gradient map as the input to the kernel estimation problem due to its simplicity and efficiency. The blurry muscle image is converted to gray scale and two derivative kernels are used to obtain the gradient maps at both the vertical and horizontal directions:
\begin{equation*}
\nabla X=\{\partial_x X,\partial_y X\}=\{X \otimes \partial_x,X \otimes \partial_y\},
\end{equation*}
where $\otimes$ denotes the convolution operation and
\begin{equation*}
\partial_x = \begin{bmatrix}
-1&1\\
0&0
\end{bmatrix},
\partial_y = \begin{bmatrix}
-1&0\\
1&0
\end{bmatrix}.
\end{equation*}

There are several options to model noise in muscle images such as Gaussain distribution, Poisson distribution \cite{Deblur_Lucy} \cite{Deblur_Richard}, and impulse distribution \cite{Deblur_noise_impulse}. For the convenience of computation, we assume the noise follows white Gaussian distribution with a uniform variance $\sigma$. The data fidelity term becomes:

\begin{equation}\label{eq:dataFidelity}
\sum^N_{i=1} E(I|x_i,k)=||\nabla X\otimes k-\nabla I||_F^2 \sim \mathcal{N}(0,\sigma),
\end{equation}
where $\nabla X=\{\partial_x x_i, \partial_y x_i\}$ and $\mathcal{N}(0,\sigma)$ denotes a white Gaussian model. Although noise is explicitly modeled and addressed, strong noise hinders the accurate estimation of the blur kernel and blur removal. We propose to use an automatic denoising framework proposed in \cite{iccv-13-xiangfei-kong} to suppress noise without the prior knowledge of the noise parameter.
\subsubsection{Image Prior}
Image priors are used as regularizations during the optimization process to alleviate the ill-posedness problem and suppress artifacts such as rings. There are large number of image priors proposed in image deblur literature. Whyte \emph{et al.} apply a simple $l_1$ norm to the Richard-Lucy deconvolution algorithm  \cite{Deblur_Whyte}. Shan \emph{et al.} adopt the fact in image statistics that the image gradient should have a heavy-tailed distribution  \cite{Deblur_ShanQi}. Rudin et al propose to minimize the total variation ($l_2$ norm) of the gradient map. Krishnan \emph{et al.} propose a normalized $l_1$ \cite{Deblur_Krishnan} to constraint the high frequency components of the latent image. More sophisticated but computationally tractable $l_0$ norm is proposed in \cite{Deblur_XuLi1}. We use $l_1$ norm on the gradient of the latent image, modeling its sparsity nature:

\begin{equation}\label{eq:imagePrior}
E(X) = ||\nabla X||_F^1.
\end{equation}
The sparsity of the gradient image is increased in blurry/defocus microscopic muscle images compared to its latent distortion free version. Thus, the $l_1$ norm can be adopted in a energy minimization framework to constrain the solutions and alleviate the ill-posedness problem in \ref{Eq:deblur_overall}. Empirically, this simple $l_1$ norm is not only effective in practice but also provides significant convenience in computation since the objective function is inherently convex. Gradient descent-like algorithms such as (fast) iterative shrinkage-thresholding algorithm (\textbf{ISTA} or \uppercase{FISTA}) \cite{ISTA} can be used to find its solution efficiently, providing we adopt Bayesian inference and assume the noise follows the white Gaussian noise model in Eq. \ref{eq:dataFidelity}. Other priors, such as the prior in \cite{Deblur_ShanQi} and the sophisticated $l_0$ norm \cite{Deblur_XuLi1}, are not used because they is computational efficient despite their promising performance. The total variation of the gradient tends to over-smooth the latent image. The normalized $l_1$ norm in \cite{Deblur_Krishnan} is highly non-convex and has many local minima, which hinders the successful estimation.  works well but it is usually slow to optimize.

\subsubsection{Kernel Prior}
Unlike typical motion blur problems in which the blur kernel can be modeled as a motion trajectory in 3D space, the degradation of the microscopic muscle image is largely the defocus problem with high variances in 2D space of the entire slide. Although we explicitly confine the blur kernel to be Gaussian at the initialization stage of the optimization, the weight assigned to this constrain is quickly decreased and diminished. We adopt $l_2$ norm to suppress the noise. $l_1$ norm tends to over-suppress the kernel in our case though it is a good regularization in common natural image deblurring problem. The kernel prior used in our method is written as:

\begin{equation}\label{eq:kernelPrior}
\begin{aligned}
        E(k) &= \eta p(k|\sigma)+\nu ||k||_F^2,\\
        p(k|\sigma) &= ||k(a,b)-\frac{1}{2\pi\sigma^2}exp\{-\frac{a^2+b^2}{2\sigma^2}\}||_F^2,
\end{aligned}
\end{equation}
where $p(k|\sigma)$ denotes the likelihood of the kernel to be a circular Gaussian with the parameter $\sigma$ which takes $(a,b)$ as  variables. $\eta$ and $\nu$ are weights of controlling the importance of both terms. $\eta$ is decayed exponentially and is set to $0$ after a few iterations.

\subsubsection{Optimization}
Combining Eq. \ref{eq:imagePrior} and Eq. \ref{eq:kernelPrior} to Eq. \ref{Eq:deblur_overall}, the overall optimization of the proposed deblur is written as:
\begin{equation}\label{eq:optm_overall}
\min_{\nabla X,k} ||\nabla X\otimes k-\nabla I||_F^2 + \lambda||\nabla X||_F^1 + \eta p(k|\sigma)+\nu ||k||_F^2,
\end{equation}
where $\lambda$, $\eta$ and $\nu$ are weight terms.

We discuss the detailed configurations of these parameters in the experiment section and the user does not have to . Eq. \ref{Eq:deblur_overall} is solved using Alternating Minimization (\uppercase{am}) and obtain the estimated blur kernel and the gradient of the latent image (which is an ``unnatural" representation of the latent image and is not used after the kernel is estimated):
\begin{equation}\label{eq:optm_img}
\min_{\nabla X} ||\nabla X\otimes k-\nabla I||_F^2 + \lambda||\nabla X||_F^1,
\end{equation}
and 

\begin{equation}\label{eq:optm_kernel}
\min_k ||\nabla X\otimes k-\nabla I||_F^2 + \eta p(k|\sigma)+\nu ||k||_F^2.
\end{equation}

To optimize Eq. \ref{eq:optm_img}, we use \uppercase{ISTA} algorithm by replacing matrix multiplications with convolutions as shown in algorithm \ref{alg:optm_latent_ISTA} and find the solution of the latent muscle image in gradient domain. Note that $\Phi_{\pi}(\cdot)$ denotes a rotation of $180^\circ$ operation on a matrix\footnote{This can be implemented via the ``rot90(k, 2)" command in \uppercase{matlab}.}. This rotation operation replaces the original matrix transpose when 1) typical \uppercase{ISTA} algorithm or 2) the Toeplitz form for convolution as in Eq. \ref{Eq:deblur_overall_v2} is used. $sign(\cdot)$ is the element-wise sign function for a matrix and $\odot$ denotes the matrix element-wise multiplication. Note that we obtain the updates for $\partial_x X$ and $\partial X_y$ individually using Eq. \ref{eq:optm_img}. In Eq. \ref{eq:optm_kernel}, the two updated gradient maps are input jointly as one term $\nabla X$.

\begin{algorithm}
	\caption{Gradient map update}
	\label{alg:optm_latent_ISTA}
	\begin{algorithmic}[1]
		\STATE \textbf{Task}: Obtain the solution of the latent image in gradient domain $\nabla \tilde{X}$ under $l_1$ norm regularization;
		\STATE \textbf{Initialization}: Observed gradient image $\nabla I$ in current resolution level, initial latent image in gradient domain $\nabla X^0\gets\nabla I$, initial kernel $k^0$ at coarsest resolution level or the optimization result $k=\tilde{k}$ from a coarser resolution level, regularization parameter $\lambda$, ISTA threshold $\zeta$, maximum inner iteration number $T$, current iteration $t \gets 1$;
		\REPEAT
		\STATE Set $\nabla Y \gets \nabla X^t-\zeta\Phi_\pi(k) \otimes (k \otimes \nabla X^t - \nabla I)$.
		\STATE Set $\nabla X^{t+1} \gets sign(\nabla Y) \odot \max(|\nabla X^t|-\lambda\zeta,0)$.
		\STATE Compute the energy cost in Eq. \ref{eq:optm_img}.
		\STATE Set $t\gets t+1$.
		\UNTIL $t>T$ or energy cost convergence reached.
		\STATE Set $\nabla \tilde{X} \gets \nabla X^t$.
	\end{algorithmic}
\end{algorithm}

\begin{algorithm}
	\caption{Kernel update}
	\label{alg:optm_Kernel}
	\begin{algorithmic}[1]
		\STATE \textbf{Task}: Obtain the solution of the blur kernel $\tilde{k}$ under both $l_2$ norm and Gaussian similarity regularization;
		\STATE \textbf{Initialization}: Observed gradient image $\nabla I$ at current resolution level,  regularization parameter $\eta$ and $\nu$ decayed by $\Delta \eta$ at each iteration;
		\STATE Compute the least square estimation of the kernel $\hat{k}$ in Eq. \ref{eq:optm_kernel_fourier}.
		\IF {$\eta\neq 0$}
		\STATE{
		\begin{itemize}
			\item[] Compute the maximum likelihood solution of the approximate Gaussian function $\mathcal{N}(0,\tilde{\sigma}) \simeq \hat{k}$ in Eq. \ref{eq:optm_kernel_gaussian}.
			\item[] Search for the solution $\tilde{k}\sim \mathcal{N}(0,\sigma+\nabla\sigma)$ in Eq. \ref{eq:optm_kernel_search}.
			\item[] Decrease the Gaussian likelihood weight $\eta \gets \max((\eta-\Delta \eta),0)$.
		\end{itemize}
	    }
	    \ELSE
	    \STATE{Set $\tilde{k} \gets \hat{k}$.}
		\ENDIF
		\STATE Normalize the estimated kernel $\tilde{k} \gets \tilde{k}/ ||\tilde{k}||_F^1$.
	\end{algorithmic}
\end{algorithm}

\begin{figure}
	\centering
	\includegraphics[width=0.8\linewidth]{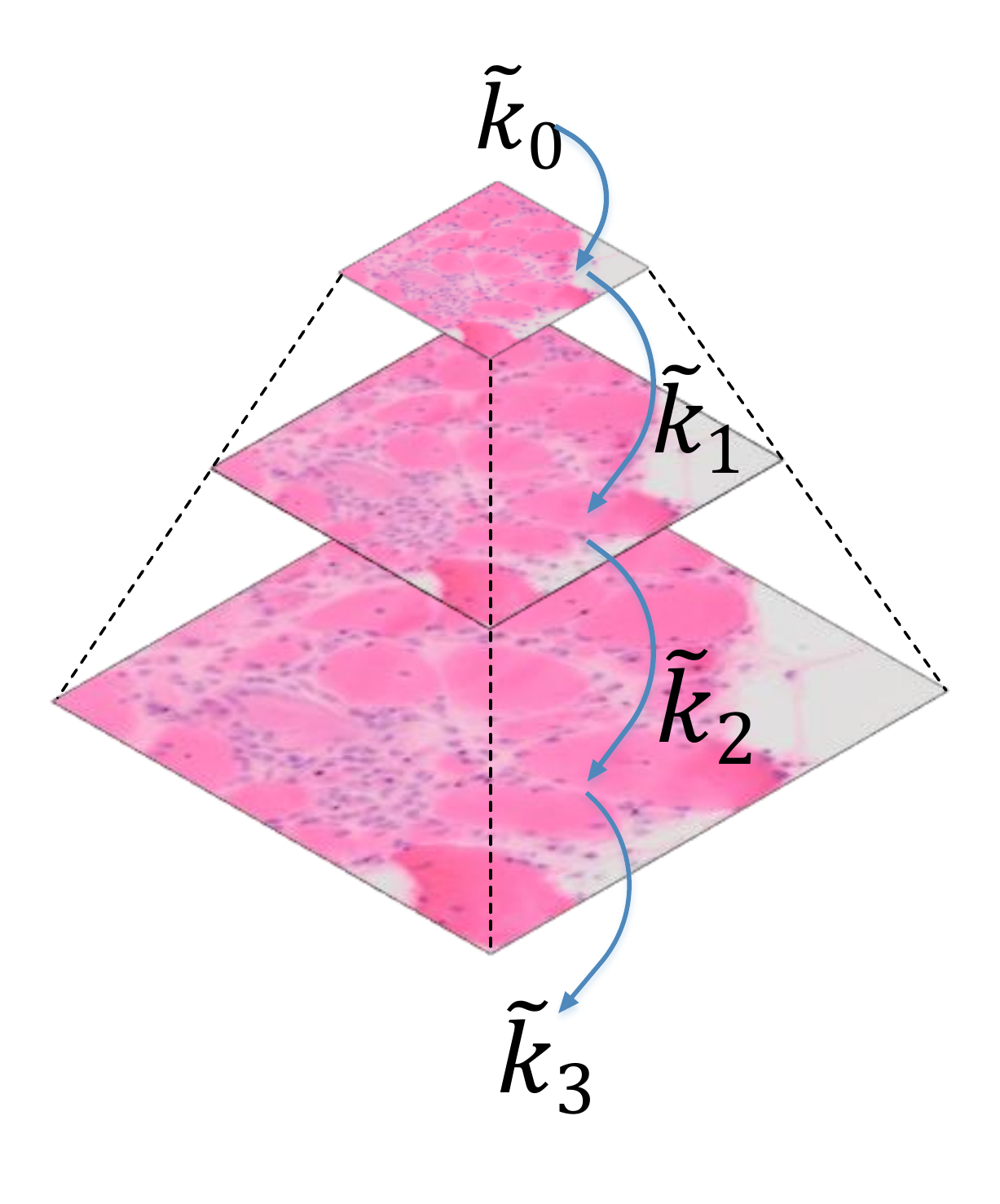}
	\caption{Pyramid representation of the muscle image. $\tilde{k}_\omega$ represents the initial/estimated blur kernels.}
	\label{Fig:pyramid}
\end{figure}

We present the details of solving the optimization problem for kernel $k$ in Eq. \ref{eq:optm_kernel}. We first solve $k$ with $l_2$ regulation in a least square minimization:

\begin{equation}
    \begin{aligned}
    \hat{k}&=\argmin_k E_k,\\
           &=\argmin_k ||\nabla X\otimes k-\nabla I||_F^2 + \nu ||k||_F^2,
    \end{aligned}
\end{equation}
whose solution, according to the Parseval's theorem\footnote{The sum of the square of a function is equal to the sum of the square of its Fourier transform.}, is efficiently obtained in closed-form by Fast Fourier Transform (\uppercase{FFT}) and setting derivative $\frac{\partial\mathcal{F}(E_k)}{\partial\mathcal{F}(k)}=0$ \cite{Deblur_Cho}:

\begin{equation}\label{eq:optm_kernel_fourier}
\begin{aligned}
    \hat{k} &= \mathcal{F}^{-1} \Big( \frac{\overline{\mathcal{F}(\nabla X)}\odot\mathcal{F}(\nabla I)}{\overline{\mathcal{F}(\nabla X)}\odot\mathcal{F}(\nabla X)+\nu \mathbf{1}} \Big)\\
            &= \mathcal{F}^{-1} \Big( \frac{\overline{\mathcal{F}(\partial_x X)}\odot\mathcal{F}(\partial_x I)+\overline{\mathcal{F}(\partial_y X)}\odot\mathcal{F}(\partial_y I)} {\overline{\mathcal{F}(\partial_x X)}\odot\mathcal{F}(\partial_x X) + \overline{\mathcal{F}(\partial_y X)}\odot\mathcal{F}(\partial_y X) + \nu \mathbf{1}} \Big),
\end{aligned}
\end{equation}
where $\mathcal{F}$ and $\mathcal{F}^{-1}$ is the Fourier/inverse Fourier transform pair\footnote{The Fourier/inverse Fourier transform pair is implemented with fast (inverse) Fourier transform (\textbf{FFT} and \textbf{iFFT}). In \uppercase{matlab}, we use ``fft2" to compute the coefficients in frequency domain for gradient maps and use ``otf2psf" to compute the inverse Fourier transform to convert the estimated blur kernel to spatial domain. This is because ``fft2" centers the input image at $(1,1)$ while ``otf2psf" centers at the geometric center.}. $\mathbf{1}$ denotes a matrix whose elements are uniformly $1$. Then, we find a Gaussian blur kernel $\mathcal{N}(0,\hat{\sigma})$ closest to $\hat{k}$ with maximum likelihood inference:

\begin{equation}\label{eq:optm_kernel_gaussian}
    \begin{aligned}
        \tilde{k}&\sim \mathcal{N}(0,\tilde{\sigma}),\\
        \tilde{\sigma}&=\argmax_{\sigma\in\Theta}\hat{\ell}(\sigma;\hat{k},\mathcal{N}(0,\tilde{\sigma})),
    \end{aligned}
\end{equation} 
where $\hat{\sigma}$ is the maximum likelihood solution of the parameter and $\Theta$ is its value domain. Finally, we search in a small window for this parameter:
\begin{equation}\label{eq:optm_kernel_search}
    \tilde{\sigma} \pm \Delta \sigma,
\end{equation}
to find a local minima for the cost function in Eq. \ref{eq:optm_kernel}. We use brute force here and search for the result for each possible value of $\tilde{\sigma} \pm \Delta \sigma_i$, where $\sigma_i$ is some preset step sizes. Eq. \ref{eq:optm_kernel_gaussian} and Eq. \ref{eq:optm_kernel_search} are served as regularizations to control the overall shape of the estimated kernel and to suppress noise. We empirically find out that they are helpful only in the first few iterations. Thus, we gradually decrease the value of the weight $\nu$ to zero after a few iterations. The brute force search in Eq. \ref{eq:optm_kernel_search} looks ad-hoc but it is fast in practice. In fact, we are not the first to propose this approach. Hu \emph{et al.} assume their estimated kernel to be plate-shaped and use brutal force to search for its axis parameters \cite{Deblur_HuZhe}. They claim that the computational cost is not a problem. In addition, we normalize the estimated kernel to keep its $l_1$ norm to be $1$ in order to preserve its energy at each iteration of the optimization. The kernel update steps are summarized in Algorithm \ref{alg:optm_Kernel}. We follow the popular coarse-to-fine strategy using pyramid representation of the muscle image as Fig. \ref{Fig:pyramid} depicts. We conduct our blur kernel estimation algorithm from low to high resolution levels. Each resolution level provides an estimation of the kernel (with an increasing size) to the next finer level. At each resolution level, we run Algorithm \ref{alg:optm_latent_ISTA} and Algorithm \ref{alg:optm_Kernel} several but not too many iterations for refinement. The overall kernel estimation algorithm is summarized in Algorithm \ref{alg:optm_overall}.

\begin{algorithm}[t]
	\caption{Overall blur kernel estimation}
	\label{alg:optm_overall}
	\begin{algorithmic}
		\STATE \textbf{Task}: Obtain the solution of the blur kernel $\tilde{k}$;
		\STATE \textbf{Initialization}: Observed image $I$, set the initial blur kernel as $k^0_w \gets\delta$ (Dirac function) or as an estimated kernel from the neighboring region $k^0_w\gets\tilde{k}'$, regularization parameter $\lambda$, $\eta$ and $\nu$, total outer iteration number $N$ and the current iteration $n$, total pyramid levels $\Omega$ and the current processing resolution level $\omega_{ini}\gets(k^0_w=\delta)?1:\Omega$, where $\omega=1$ relates to the coarsest resolution level of the image;
		\STATE Compute the image pyramid $\{I_\omega\}$ with $\Omega$ levels for $I$.
		\FOR {$\omega = \omega_{ini}$ to $\Omega$}
		\STATE{ 
			Initialize the gradient map at the current resolution level: $\nabla X^0_\omega \gets (\omega = 1)? \nabla I_\omega:\nabla \tilde{X}_{\omega-1}$.
			
			Set $\nu$ as its initial value.
			\FOR {$n \gets 1$ to $N$}
			\STATE{
				\begin{itemize}
					\item[] Update the blur kernel $\tilde{k}^n_\omega$ for $\nabla I_\omega$ at the current resolution level with Algorithm \ref{alg:optm_Kernel} by inputting $\nabla X = \nabla X^n_\omega$.
					\item[] Update the gradient map $\nabla X^n_\omega $ of the latent image at the current resolution level with Algorithm \ref{alg:optm_latent_ISTA} by inputting $k=k^n_\omega$.
				\end{itemize}
			}
			\ENDFOR}
		\STATE{
			\IF{$\omega \neq \Omega$}
			    \STATE{Upscale $\tilde{k}_\omega$ and $\nabla \tilde{X}_\omega$ to the next finer resolution level.}
			\ENDIF
			}
		\ENDFOR
	\end{algorithmic}
\end{algorithm}

\subsection{Latent Image Recovery}
After the kernel for the finest resolution level is estimated, we deconvolve the muscle image with it and obtain the estimation of the latent image. Note that we only obtain the latent image in gradient domain in previous steps. Numerous works for non-blind (with knowledge of the kernel) deconvolution have been proposed in literature, ranging from the traditional inverse filtering and Weiner filtering \cite{WeinerFilter} to the modern sophisticated approaches focusing on artifact suppression, e.g., Laplacian regularization \cite{Deconvolution_Levin}, $l_1$ regularization on Richard-Lucy algorithm \cite{Deblur_Whyte} \cite{Deconvolution_l1RL_MBrown}, expectation maximization (\uppercase{em}) scheme with outliers (saturated pixels and noise) handling \cite{Deconvolution_Cho}, or a hybrid method combining the advantages from the previous two \cite{Deblur_HuZhe}. We adopt the Laplacian regularization based approach in \cite{Deconvolution_Krishnan} to deconvolve the degraded muscle images with estimated blur kernels due to its good performance and fast convergence:

\begin{equation}\label{eq:deconvolution}
\tilde{X} = \argmin_{X}\beta||X\otimes \tilde{k}-I||^2_F+||\nabla X||_{\alpha},
\end{equation}
where $\tilde{X}$ is the enhanced image, $\tilde{k}$ is the estimated blur kernel from Eq. \ref{eq:optm_overall}. We set parameters such as $\beta=3000$ and $\alpha=0.8$ exactly the same as suggested in \cite{Deblur_Krishnan}.

\subsection{Local Kernel Propagation}
\begin{figure*}
	\begin{center}
		\begin{tabular}{c}
            \includegraphics[width=\swidthone]{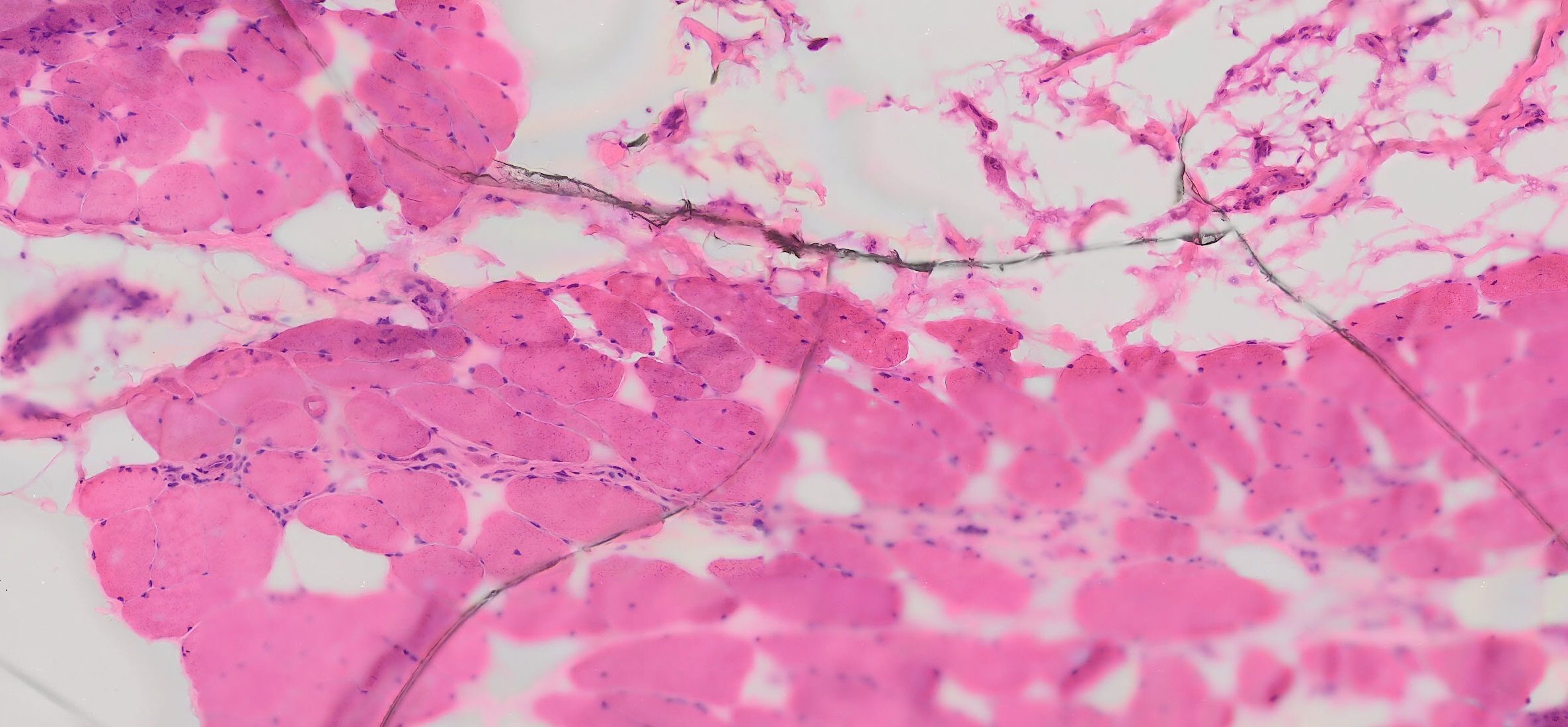} \\ \includegraphics[width=\swidthone]{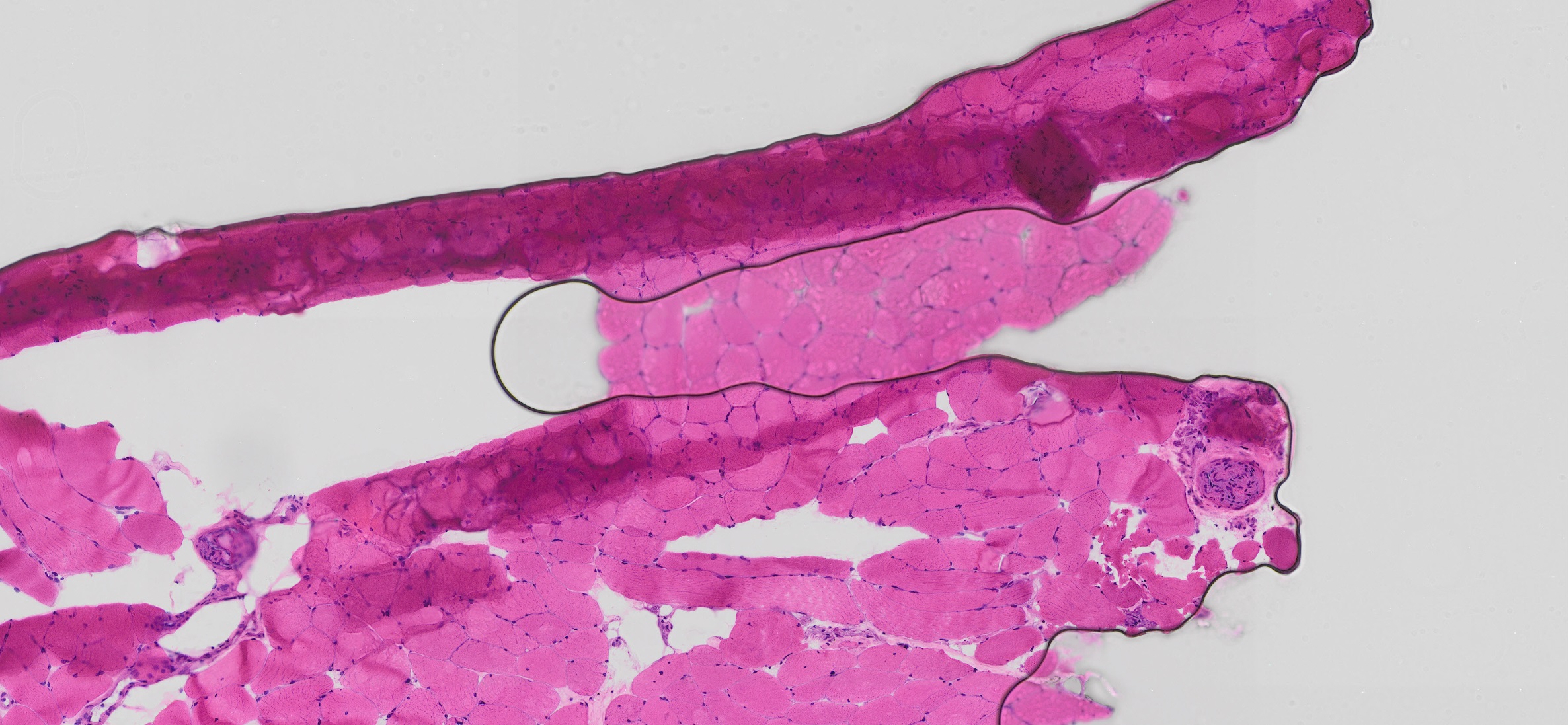}
		\end{tabular}
		\caption{Examples of the non-uniform blur on the muscle images.}
		\label{Fig:non-uniform}
	\end{center}
\end{figure*}

The proposed deblur algorithm until now assumes the blur kernel is constant across the image. This is not true on typical microscopic muscle images where blur and defocus happen in a very non-uniform style. See Fig. \ref{Fig:non-uniform} for examples. Rather than on the entire scope, we perform our deblur algorithms on muscle sub-image (200 by 200 pixels in this paper) and optimize the local gradient sparsity in Eq. \ref{eq:imagePrior} of the patch and obtain a group of spatially variant blur kernels. The entire image is thus recovered via the non-blind deconvolution on each of the sub-images using Eq. \ref{eq:deconvolution}. The recovered sub-images are finally stitched. We extract sub-images with small marginal overlaps. The recovered sub-image are simply averaged on the overlapped parts to preserve inter sub-image coherence of the image structure. Although this method is simple, it benefits the blur removal of the microscopic muscle image in several ways:
\begin{enumerate}
	\item The strong non-uniform nature of the blur kernel of the muscle image is addressed. In the experiment section, we show that the proposed method outperforms the existing non-uniform deblur algorithms significantly due to this flexibility.
	\item Fast deblur computations is enabled by feeding the cpu/gpu with image patches in parallel. The deblur algorithm is typically very slow on the entire, large muscle image.
\end{enumerate}

\begin{figure}
	\begin{center}
		\includegraphics[width=\linewidth]{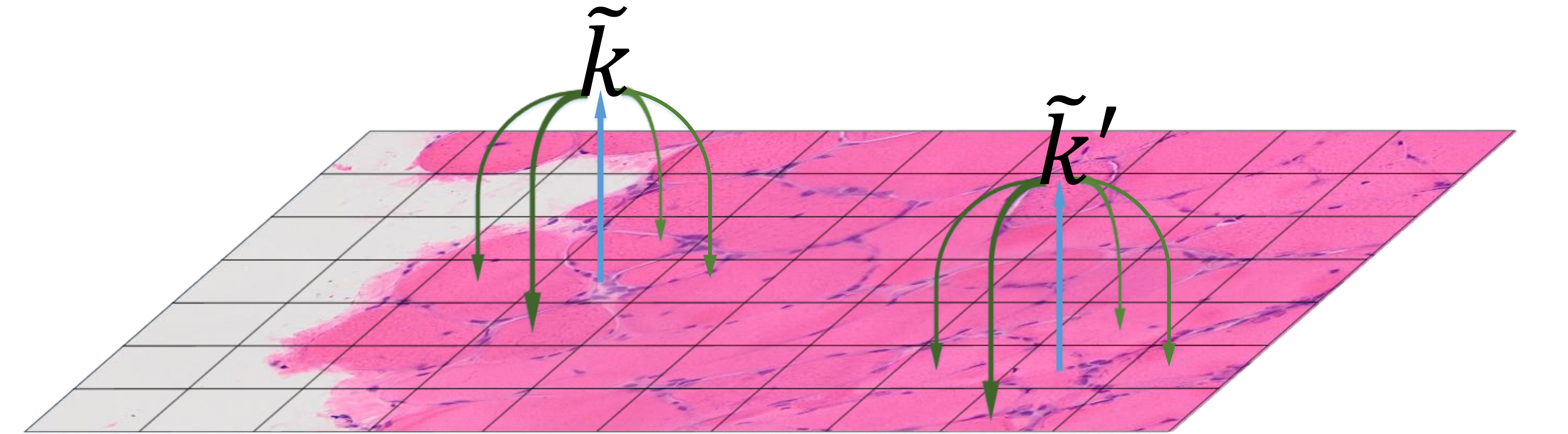}
		\caption{Local kernel propagation. The estimated blur kernels $\tilde{k}$ and $\tilde{k}'$ are propagated to their neighboring subregions to accelerate their kernel estimations.}
		\label{Fig:propagation}
	\end{center}
\end{figure}

To further speedup the algorithm, we randomly randomly pick up $n$ sub-image centers\footnote{We set $n$ as the number of the \uppercase{cpu}s of our experiment computer.} away from the boundary (to avoid picking sub-images that have too few number of neighbors). We run Algorithm \ref{alg:optm_overall} and estimate their blur kernels individually. The estimated kernels are treated as seeds and are propagated to their neighboring sub-images as initializations during their kernel estimation. The rationale behind this method is that most neighboring sub-image share similar blur degree and thus should have similar blur kernels (Globally, they still display high non-uniformity). The sub-images having their kernels propagated by seeds are not represented with pyramid structure during the kernel estimation\footnote{See the different initialization of of $\omega_{ini}$ for seeds and non-seeds in Algorithm \ref{alg:optm_overall}.} Thus, the computation is accelerated by skipping the kernel estimation on coarser resolution levels. Fig. \ref{Fig:propagation} shows an example of kernel propagation made by to estimated kernels.



\section{Experimental Results}
\label{sec:experiment}
\subsection{Experiment Settings}
We obtain $44$ microscopic muscle images using an Olympus VS 120 advanced optical microscope, each of which has very large size. Thus, the muscle images used in our experiments are cropped with large size, assuming each channel of the muscle image data is normalized to range in $[0,1]$. We empirically set $\lambda=80$, $\eta=15$, $\nu=6$ and decay term $\Delta \nu=2$ and do not make any further tunings. We assume the kernel size to be $15$ pixel\footnote{This size is the maximum size while in practice most estimated blur kernels are smaller than it.}. The outer iteration number $N$ and the inner iteration number $T$ are set as $5$ and $3$, respectively. The search window in Eq. \ref{eq:optm_kernel_search} is set as $\Delta \sigma = \{0.1/255,0.2/255,...0.5/255\}$. The number of resolution levels of the image pyramid $\Omega$ is set as $5$.  We test the proposed framework on a desktop computer with Intel Xeon E5-1650 3.50 GHz (12 threads) \uppercase{cpu} and 128GB memory.
\subsection{Synthesis Blur}
\begin{figure}
	\begin{center}
	    \includegraphics[width=0.7\linewidth]{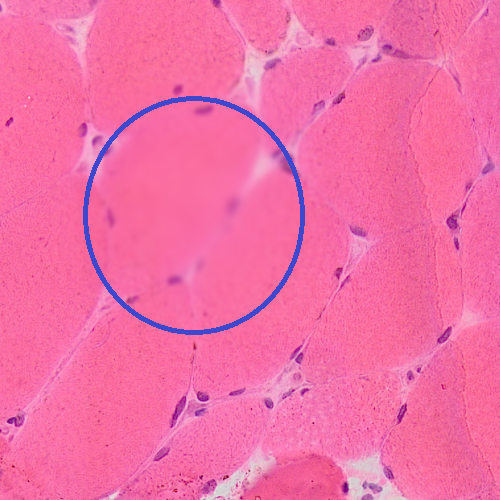}
    \end{center}
    \caption{An example of the muscle image corrupted with the synthesis blur. Blur circle locates roughly the blurry area.}
    \label{fig:synthesis_blur_example}
\end{figure}

In this experiment, we manually crop $20$ muscle sub-images (round $2000$ by $1000$)\footnote{The actual size of the } that we consider as sharp ones without visually obvious blur or defocus. We synthesize a number of Gaussian blur on them to simulate the spatially variant blur and defocus. We denote $b_i=\{(a_i,b_i),\sigma_s,\sigma_l\}$ as a blur of the latent muscle image centered at the coordinate $(a_i,b_i)$ and compute the blurred image as follows:
\begin{equation}
    I(m,n) = \sum_{(i,j)\in\Phi(m,n)} \frac{1}{W_k}\cdot X(i,j) \cdot k(i,j),
\end{equation}
where $I(m,n)$ is the pixel value of the blurred muscle image, $(i,j)$ denotes the arbitrary location in the neighborhood of the central pixel $\Phi(m,n)$, $X$ is the sharp muscle image and $k$ is the synthesis Gaussian blur kernel with its normalization term $W_k$ defined as:
\begin{equation*}
\begin{split}
    k &\sim \mathcal{N}(0,\rho(m,n)),\\
    W_k&=\sum_{(x,y)\in k} k(x,y),
\end{split}
\end{equation*}
where $\rho(m,n)$ is the standard deviation of the blur kernel. It is determined by the distance between $(m,n)$ and the blur center $(a,b)$:
\begin{equation*}
    \rho(m,n) = \sigma_s \cdot exp\{\frac{-|(a,b)-(m,n)|^2}{2\sigma_l}\},
\end{equation*}
where $\sigma_s$ is the blur strength of the kernel $b_i$ and is decayed as the location deviates from the blur center and is controlled by the parameter $\sigma_l$. We randomly set the center of the blur kernel and set $\sigma_s\in (0.5/255,4.5/255]$, $\sigma_l=100$. Figure \ref{fig:synthesis_blur_example} shows a distortion free muscle image corrupted by the defined synthesis blur with $\sigma_s=5/255$. We uniformly split the clean images into four distortion levels degraded by different number of synthesis Gaussian kernels and different $\sigma_s$. The settings of the distortion levels are described in Table \ref{tab:distortion}.
\begin{table*}
	\centering
	\caption{Settings of different levels of synthesis distortion.}
	\begin{tabular}{|c|c|c|}\hline
		distortion level & number of Gaussian blurs & $\sigma_s$\\ \hline
		Level I   & 5  & $\in$(0.5,1.5]\\ \hline
		Level II  & 10 & $\in$(1.5,2.5]\\ \hline
		Level III & 15 & $\in$(2.5,3.5]\\ \hline
		Level IV  & 20 & $\in$(3.5,4.5]\\ \hline	 
	\end{tabular}
	\label{tab:distortion}
\end{table*}
At each level, we randomly choose the value of $\sigma_s$ uniformly distributed at the interval described in \ref{tab:distortion}, whose value is for pixel value range $[0,255]$ and is dived by $255$ in our experiment.

We test the proposed deblur algorithm of our framework and set parameters as mentioned before without further tweaking. We compare our the visual quality of our results with the state-of-art spatially non-uniform deblur algorithm by L. Xu \emph{et al.} \cite{Deblur_XuLi1} using peak signal to noise ratio (\textbf{PSNR}) as the image quality metric. The 

\begin{table*}[htbp]
	\centering
	\caption{Comparison of the enhancement results for synthesis blur. Average PSNR value is compared at each distortion level.}
	\begin{tabular}{|c|c|c|c|c|} \hline
		& Level I & Level II & Level III & Level IV \\ \hline
		L. Xu\cite{Deblur_XuLi1} & 42.34 & 36.56 & 30.10 & 23.65\\ \hline
		proposed & 47.64 & 41.26 & 32.13 & 25.02 \\ \hline
	\end{tabular}%
	\label{tab:exp_synthesis}%
\end{table*}%

\begin{table*}[htbp]
	\centering
	\caption{Comparison of the enhancement results for real blur using blind IQA score.}
	\begin{tabular}{|c|c|c|c|} \hline
		Image $\#$. & IQA score & IQA score (proposed) & IQA score (\cite{Deblur_XuLi1}) \\ \hline
		01 & 0.28 & \textbf{0.44} & 0.37 \\ \hline
		02 & 0.43 & \textbf{0.49} & 0.43 \\ \hline
		03 & 0.54 & \textbf{0.57} & 0.56 \\ \hline
		04 & 0.14 & \textbf{0.39} & 0.20 \\ \hline
		05 & 0.58 & 0.53 & \textbf{0.56} \\ \hline
		06 & 0.38 & 0.40 & \textbf{0.42} \\ \hline
		07 & 0.66 & \textbf{0.6}9 & 0.67 \\ \hline
		08 & 0.40 & \textbf{0.77} & 0.47 \\ \hline
		09 & 0.34 & \textbf{0.66} & 0.39 \\ \hline
		10 & 0.43 & \textbf{0.64} & 0.46 \\ \hline
		11 & 0.44 & 0.43 & \textbf{0.48} \\ \hline
		12 & 0.47 & \textbf{0.65} & 0.46 \\ \hline
	\end{tabular}%
	\label{tab:exp_real}%
\end{table*}%
\subsection{Real Blur}
We test our algorithm on 12 cropped muscle images (around $2000$ by $1000$) with significant noticeable blur. To quantitatively evaluate the performance, we use a popular blind image quality assessment (blind \emph{IQA}) \cite{Quality} to compare the results with \cite{Deblur_XuLi1}. The blind IQA in \cite{Quality} is a image quality evaluation scoring system which returns a score regarding to the image's sharpness, noise, transmission loss, etc. without the necessity of the availability of the distortion-free version of the image. We normalize the score to the range [0,1] for convenience. Table \ref{tab:exp_real} summarizes the blind IQA scores for the original blur images, enhanced by the proposed algorithm, and those by \cite{Deblur_XuLi1}. It is observed that the proposed algorithm outperforms the \cite{Deblur_XuLi1} on 9 out of 12 blur muscle images, although \cite{Deblur_XuLi1} exhibits a little more robustness on certain images. The parameters for both compared methods are set with consistent configurations we empirically choose and defaulted by the authors, respectively. 

\begin{figure}
	\begin{center}
		\includegraphics[width=1\linewidth]{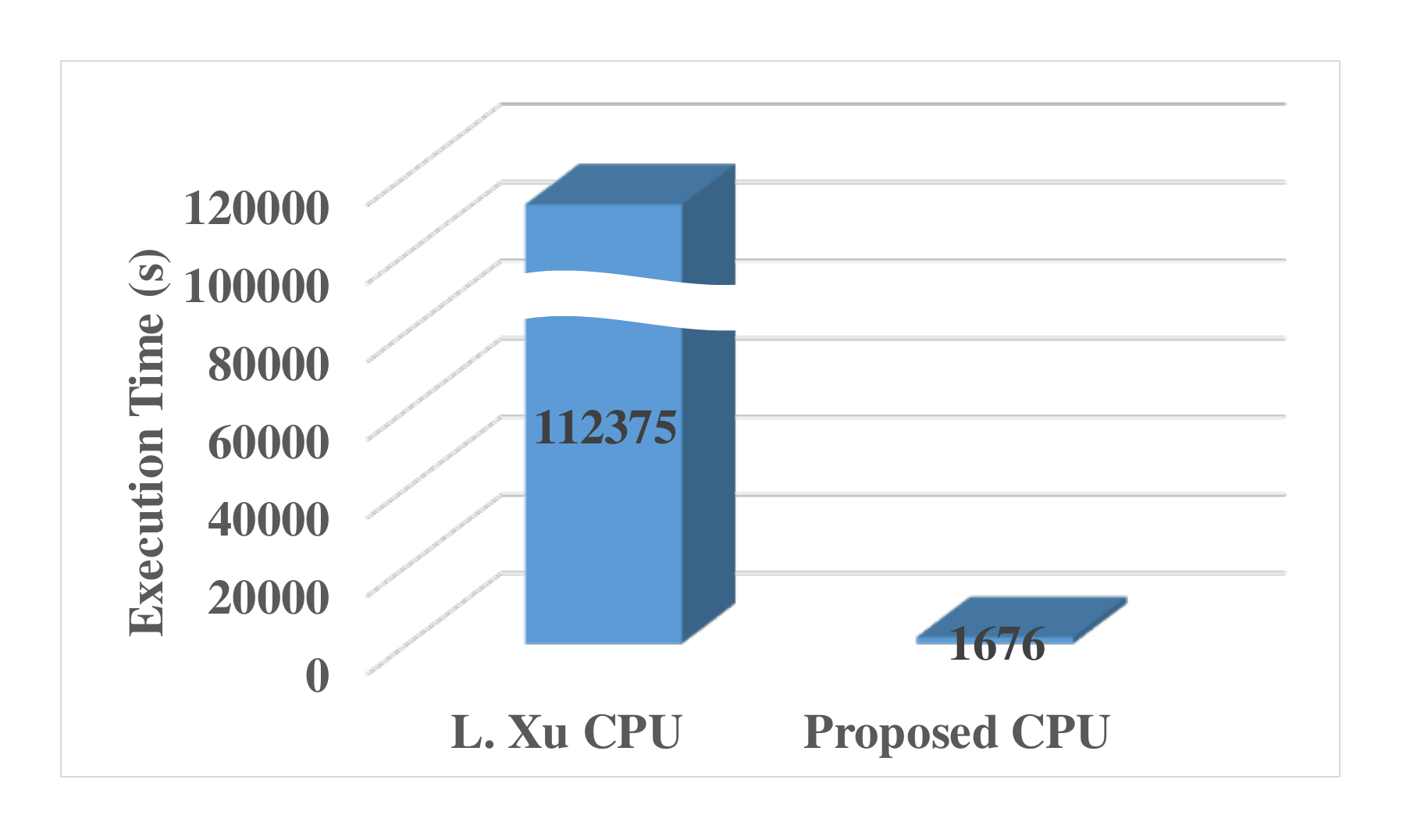}
	\end{center}
	\caption{Average execution time per one cropped muscle image for the proposed algorithm and \cite{Deblur_XuLi1}.}
	\label{fig:time}
\end{figure}

The proposed algorithm is also computationally efficient due to its simple design. The average computation time for one cropped muscle image of its CPU implementation on Matlab, compared with \cite{Deblur_XuLi1}, is presented in figure \ref{fig:time}. The proposed algorithm is significantly efficient compared with the algorithm in \cite{Deblur_XuLi1}, since the latter is designed for natural image deblur which is typically not so large as muscle images are.

\begin{figure*}
	\begin{center}
		\begin{tabular}{cc}
			\includegraphics[width=\swidthtwo]{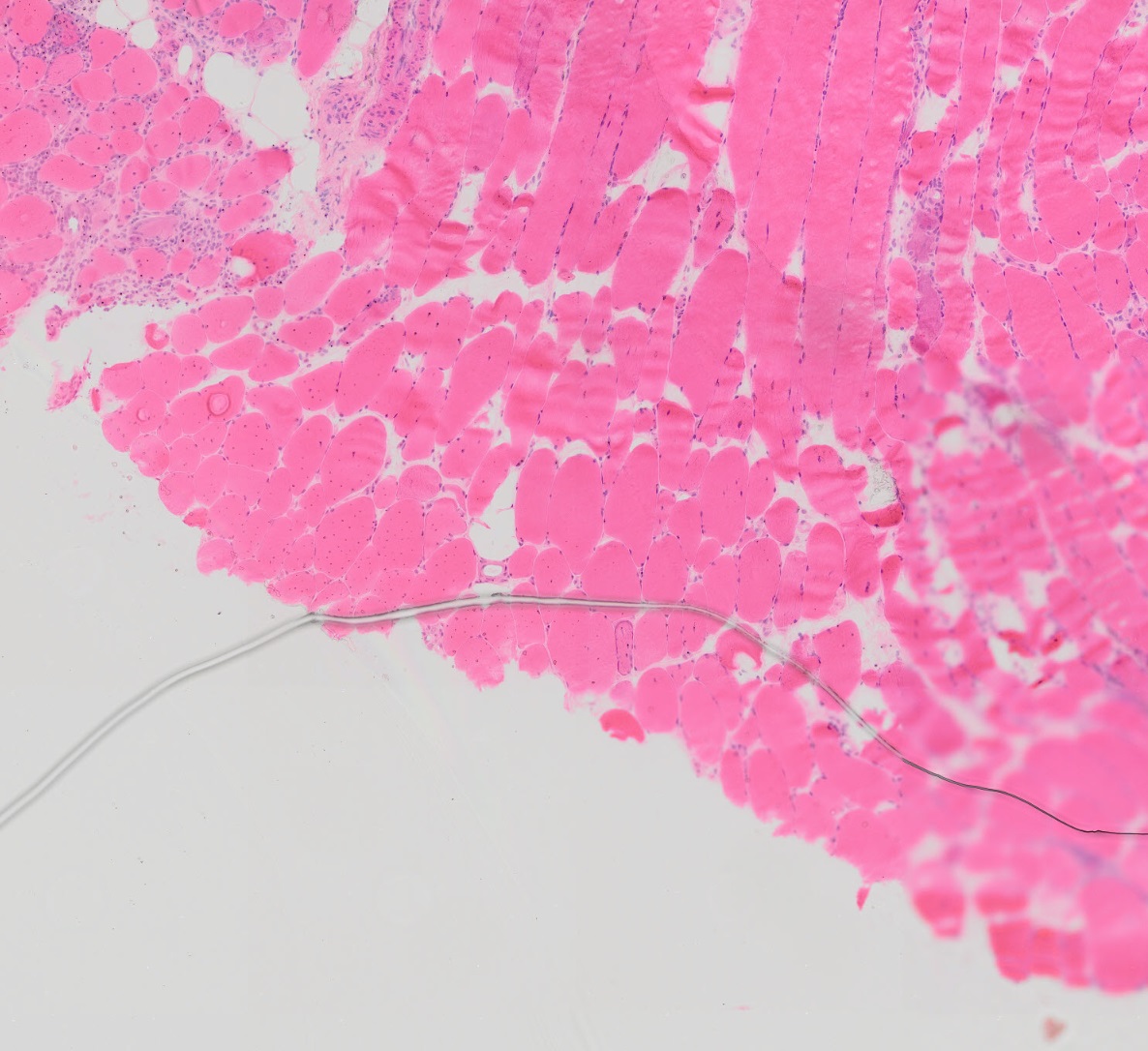} &
			\includegraphics[width=\swidthtwo]{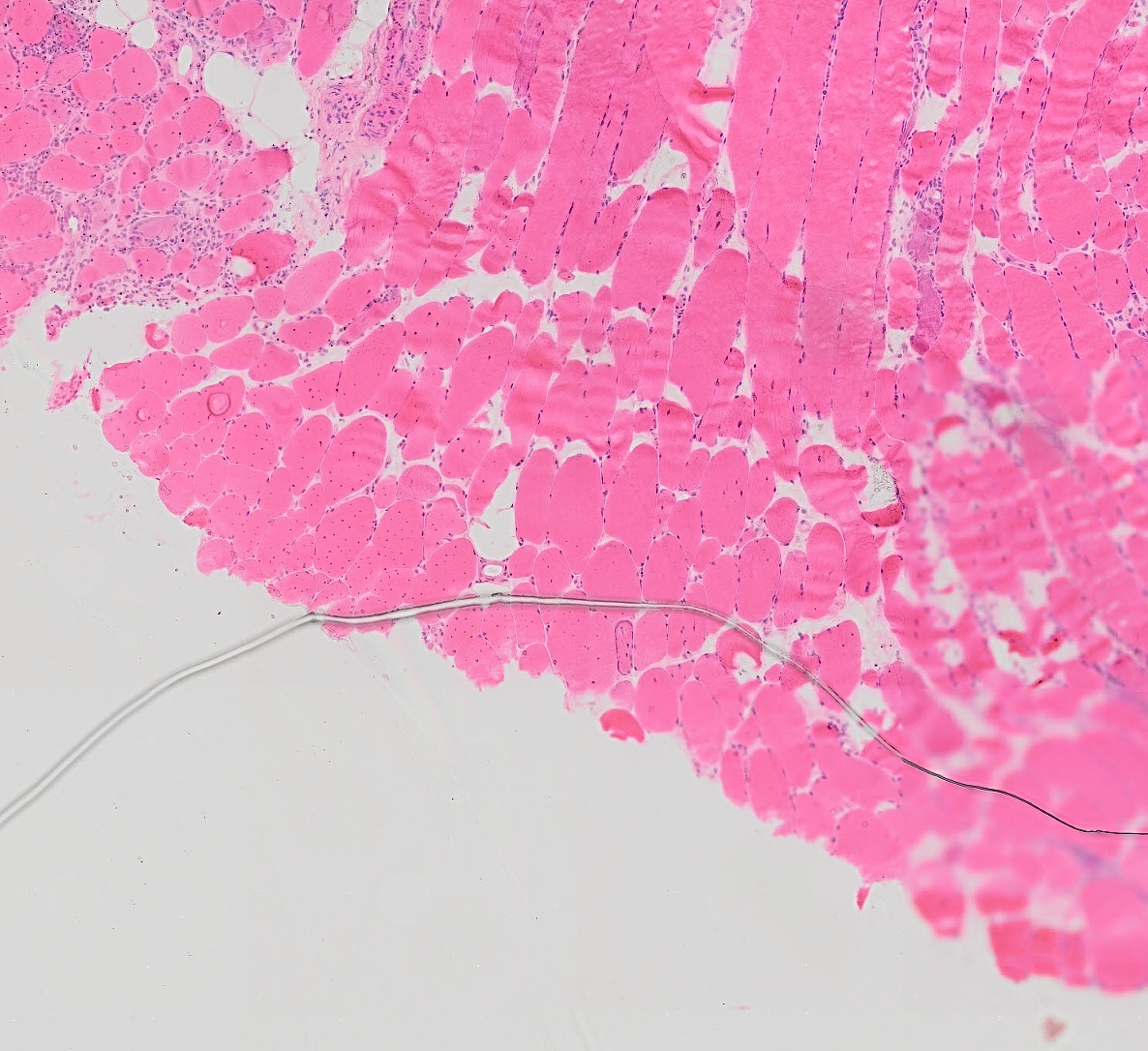}\\
			\includegraphics[width=\swidthtwo]{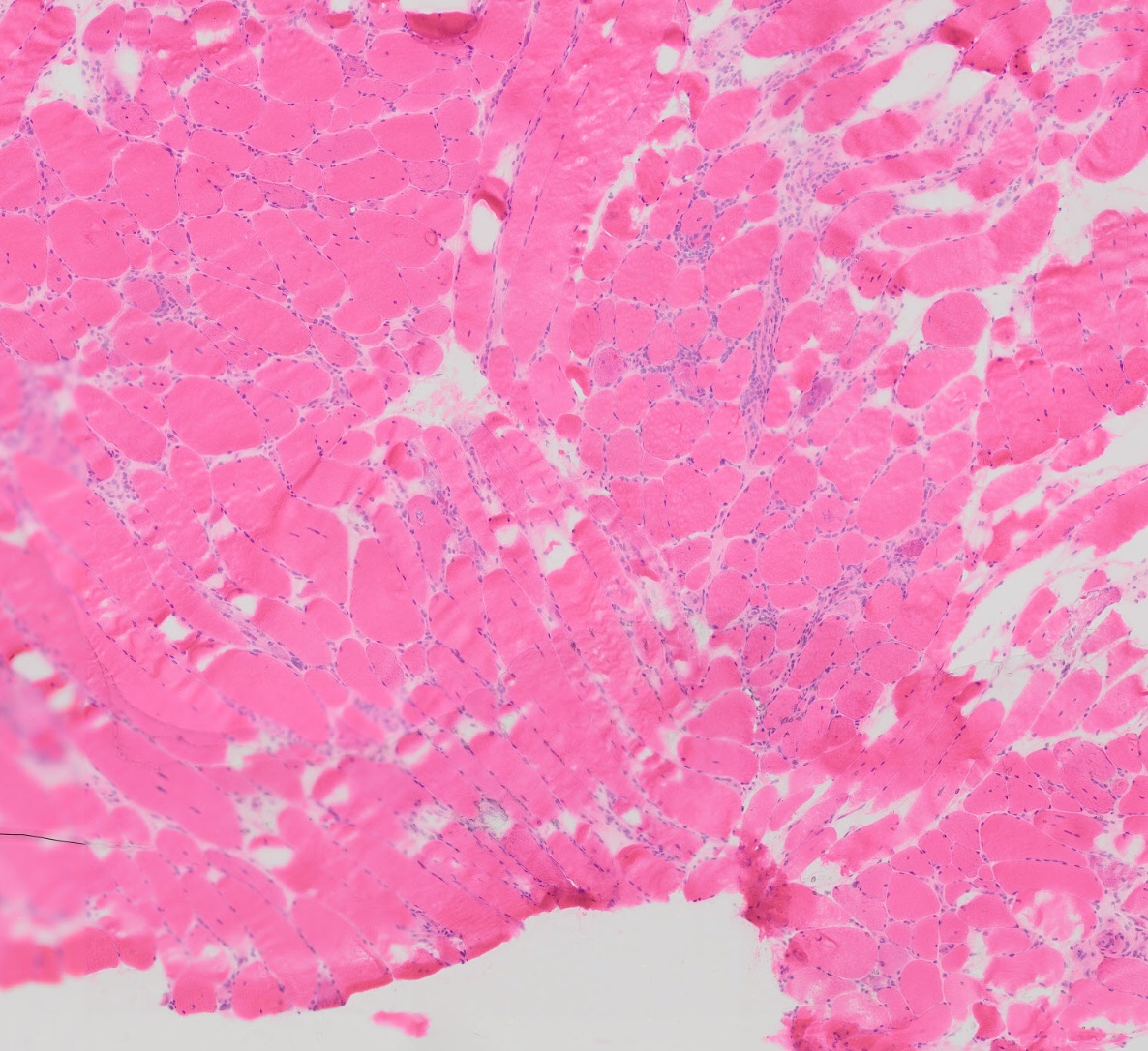} & \includegraphics[width=\swidthtwo]{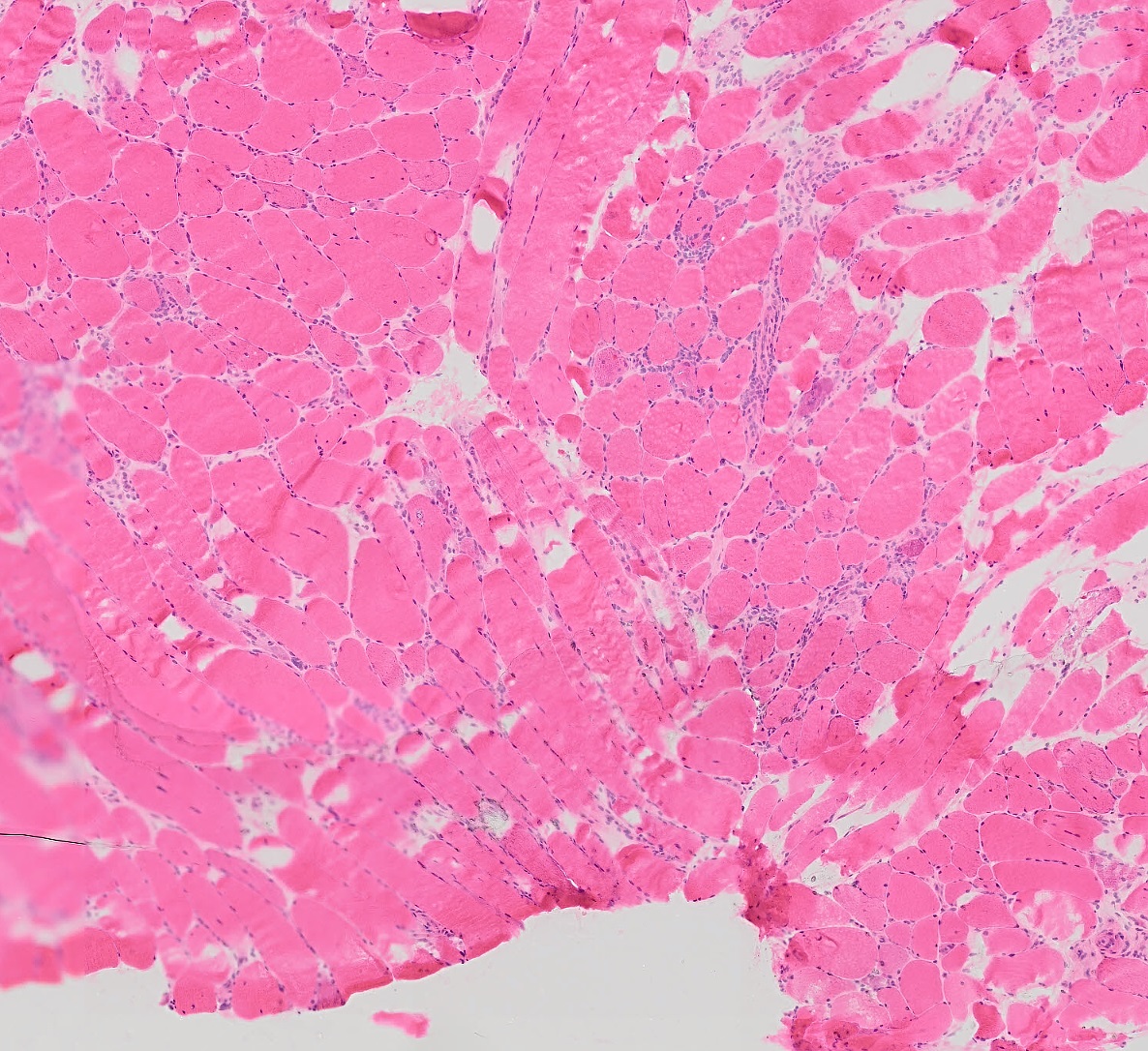}\\
		\end{tabular}
	\end{center}
	\caption{Examples of enhanced muscle images by the proposed algorithm. The figures in the left column are the original blur image and the enhanced images are in the right column. Best reviewed in enlarged resolution.}
	\label{Fig:occlusion}
\end{figure*}

\section{Conclusion}
\label{sec:conclusion}
We present a novel image enhance system in which a simple but effective deblur algorithm specialized for muscle image is involved. The system is designed with one-click-style interface which frees the users from laborious parameter tunings. The effectiveness of the algorithm is verified on both synthesis and practical blur quantified by an objective image quality evaluation metric. 

\section*{Acknowledgements}
The authors would like to thank to XX projects, fundings, and institutions.
\bibliographystyle{IEEEtran}
\bibliography{IEEEabrv,ref}

\begin{thebibliography}{10}
\providecommand{\url}[1]{#1}
\csname url@samestyle\endcsname
\providecommand{\newblock}{\relax}
\providecommand{\bibinfo}[2]{#2}
\providecommand{\BIBentrySTDinterwordspacing}{\spaceskip=0pt\relax}
\providecommand{\BIBentryALTinterwordstretchfactor}{4}
\providecommand{\BIBentryALTinterwordspacing}{\spaceskip=\fontdimen2\font plus
\BIBentryALTinterwordstretchfactor\fontdimen3\font minus
  \fontdimen4\font\relax}
\providecommand{\BIBforeignlanguage}[2]{{%
\expandafter\ifx\csname l@#1\endcsname\relax
\typeout{** WARNING: IEEEtran.bst: No hyphenation pattern has been}%
\typeout{** loaded for the language `#1'. Using the pattern for}%
\typeout{** the default language instead.}%
\else
\language=\csname l@#1\endcsname
\fi
#2}}
\providecommand{\BIBdecl}{\relax}
\BIBdecl

\bibitem{Muscle_fiber}
G.~McCall, W.~Byrnes, A.~Dickinson, P.~Pattany, and S.~Fleck, ``Muscle fiber
  hypertrophy, hyperplasia, and capillary density in college men after
  resistance training,'' \emph{Journal of applied physiology}, vol.~81, no.~5,
  pp. 2004--2012, 1996.

\bibitem{Muscle_strength1}
R.~Cooke, ``Actomyosin interaction in striated muscle,'' \emph{Physiological
  Reviews}, vol.~77, no.~3, pp. 671--697, 1997.

\bibitem{Muscle_strength2}
T.~Moritani \emph{et~al.}, ``Neural factors versus hypertrophy in the time
  course of muscle strength gain.'' \emph{American Journal of Physical Medicine
  \& Rehabilitation}, vol.~58, no.~3, pp. 115--130, 1979.

\bibitem{Muscle_IIM}
M.~C. Dalakas, ``Muscle biopsy findings in inflammatory myopathies,''
  \emph{Rheumatic Disease Clinics of North America}, vol.~28, no.~4, pp.
  779--798, 2002.

\bibitem{Aberration_nature}
M.~Haider, S.~Uhlemann, E.~Schwan, H.~Rose, B.~Kabius, and K.~Urban, ``Electron
  microscopy image enhanced,'' \emph{Nature}, vol. 392, pp. 768--769, Apr.
  1998.

\bibitem{WeinerFilter}
N.~Wiener, ``Extrapolation, interpolation, and smoothing of stationary time
  series,'' \emph{MIT Press}, 1964.

\bibitem{Deblur_Lucy}
L.~Lucy, ``An iterative technique for the rectification of observed
  distributions,'' \emph{Astronomical Journal}, vol.~79, no.~6, pp. 745--754,
  1974.

\bibitem{Defocus}
Y.-W. Tai, H.~Tang, M.~S. Brown, and S.~Lin, ``Detail recovery for single-image
  defocus blur,'' \emph{IPSJ Transactions on Computer Vision and Applications},
  vol.~1, pp. 95--104, 2009.

\bibitem{imageStatistics}
Y.~Weiss and W.~Freeman, ``What makes a good model of natural images?'' in
  \emph{Computer Vision and Pattern Recognition, 2007. CVPR '07. IEEE
  Conference on}, June 2007, pp. 1--8.

\bibitem{Deblur_ShanQi}
Q.~Shan, J.~Jia, and A.~Agarwala, ``High-quality motion deblurring from a
  single image,'' \emph{ACM Trans. Graph.}, vol.~27, no.~3, pp. 73:1--73:10,
  2008.

\bibitem{Deblur_PowerSpectrum1}
A.~Goldstein and R.~Fattal, ``Blur-kernel estimation from spectral
  irregularities,'' in \emph{Computer Vision – ECCV 2012}, 2012, vol. 7576,
  pp. 622--635.

\bibitem{Deblur_PowerSpectrum2}
T.~Yue, S.~Cho, J.~Wang, and Q.~Dai, ``Hybrid image deblurring by fusing edge
  and power spectrum information,'' in \emph{Computer Vision – ECCV 2014},
  2014, vol. 8695, pp. 79--93.

\bibitem{Deblur_Whyte}
O.~Whyte, J.~Sivic, and A.~Zisserman, ``\BIBforeignlanguage{English}{Deblurring
  shaken and partially saturated images},''
  \emph{\BIBforeignlanguage{English}{International Journal of Computer
  Vision}}, vol. 110, no.~2, pp. 185--201, 2014.

\bibitem{Deblur_Krishnan}
D.~Krishnan, T.~Tay, and R.~Fergus, ``Blind deconvolution using a normalized
  sparsity measure,'' in \emph{Computer Vision and Pattern Recognition (CVPR),
  2011 IEEE Conference on}, June 2011, pp. 233--240.

\bibitem{Deblur_XuLi2}
L.~Xu and J.~Jia, ``Two-phase kernel estimation for robust motion deblurring,''
  in \emph{Computer Vision – ECCV 2010}, vol. 6311, 2010, pp. 157--170.

\bibitem{Deblur_XuLi_l0}
L.~Xu, C.~Lu, Y.~Xu, and J.~Jia, ``Image smoothing via l0 gradient
  minimization,'' \emph{ACM Trans. Graph.}, vol.~30, no.~6, pp. 174:1--174:12,
  2011.

\bibitem{Deblur_XuLi1}
L.~Xu, S.~Zheng, and J.~Jia, ``Unnatural \textbf{L}0 sparse representation for
  natural image deblurring,'' in \emph{Computer Vision and Pattern Recognition
  (CVPR), 2013 IEEE Conference on}, June 2013, pp. 1107--1114.

\bibitem{Deblur_HuZhe}
Z.~Hu, S.~Cho, J.~Wang, and M.-H. Yang, ``Deblurring low-light images with
  light streaks,'' in \emph{Computer Vision and Pattern Recognition (CVPR),
  2014 IEEE Conference on}, June 2014, pp. 3382--3389.

\bibitem{Deblur_Cho}
S.~Cho and S.~Lee, ``Fast motion deblurring,'' \emph{ACM Trans. Graph.},
  vol.~28, no.~5, pp. 145:1--145:8, 2009.

\bibitem{Deconvolution_Cho}
S.~Cho, J.~Wang, and S.~Lee, ``Handling outliers in non-blind image
  deconvolution,'' in \emph{Computer Vision (ICCV), 2011 IEEE International
  Conference on}, Nov 2011, pp. 495--502.

\bibitem{Deblur_benckmark}
R.~Köhler, M.~Hirsch, B.~Mohler, B.~Schölkopf, and S.~Harmeling, ``Recording
  and playback of camera shake: Benchmarking blind deconvolution with a
  real-world database,'' in \emph{Computer Vision – ECCV 2012}, 2012, pp.
  27--40.

\bibitem{Deblur_nonUniform_Gupa}
A.~Gupta, N.~Joshi, C.~Lawrence~Zitnick, M.~Cohen, and B.~Curless, ``Single
  image deblurring using motion density functions,'' in \emph{Computer Vision
  – ECCV 2010}, 2010, pp. 171--184.

\bibitem{Deconvolution_l1RL_MBrown}
Y.-W. Tai, P.~Tan, and M.~S. Brown, ``Richardson-lucy deblurring for scenes
  under a projective motion path,'' \emph{IEEE Transactions on Pattern Analysis
  and Machine Intelligence}, vol.~33, no.~8, pp. 1603--1618, 2011.

\bibitem{Deblur_Richard}
W.~Richardson, ``Bayesian-based iterative method of image restoration,''
  \emph{Journal of the Optical Society of America}, vol.~62, no.~1, pp. 55--59,
  Jan. 1972.

\bibitem{Deblur_noise_impulse}
L.~Bar, N.~Kiryati, and N.~Sochen, ``Image deblurring in the presence of
  impulsive noise,'' \emph{International Journal of Computer Vision}, vol.~70,
  no.~3, pp. 279--298, 2006.

\bibitem{iccv-13-xiangfei-kong}
X.~Kong, K.~Li, Q.~Yang, W.~Liu, and M.-H. Yang, ``A new image quality metric
  for image auto-denoising,'' in \emph{ICCV}, 2013, pp. 2888--2895.

\bibitem{ISTA}
A.~Beck and M.~Teboulle, ``A fast iterative shrinkage-thresholding algorithm
  for linear inverse problems,'' \emph{SIAM Journal on Imaging Sciences},
  vol.~2, no.~1, pp. 183--202, 2009.

\bibitem{Deconvolution_Levin}
A.~Levin, R.~Fergus, F.~Durand, and W.~T. Freeman, ``Image and depth from a
  conventional camera with a coded aperture,'' \emph{ACM Trans. Graph.},
  vol.~26, no.~3, Jul. 2007.

\bibitem{Deconvolution_Krishnan}
D.~Krishnan and R.~Fergus, ``Fast image deconvolution using hyper-laplacian
  priors,'' in \emph{Advances in Neural Information Processing Systems 22},
  2009, pp. 1033--1041.

\bibitem{Quality}
M.~Saad, A.~Bovik, and C.~Charrier, ``Blind image quality assessment: A natural
  scene statistics approach in the dct domain,'' \emph{Image Processing, IEEE
  Transactions on}, vol.~21, no.~8, pp. 3339--3352, Aug 2012.

\end{thebibliography}

\end{document}